\documentclass[11pt,a4paper]{article}
\usepackage[hyperref]{acl2017}
\usepackage{times}
\usepackage{url}
\usepackage{latexsym}
\usepackage{bm}
\usepackage{graphicx}
\usepackage{amsmath}
\usepackage{amssymb}
\usepackage{harpoon}
\usepackage{array}
\usepackage{color}

\usepackage{amsfonts, amssymb, amsmath}
\usepackage{algorithm, algorithmic, amsthm}
\usepackage{graphicx}
\usepackage{color}
\usepackage{hyperref}
\usepackage[top=2.5cm, bottom=3cm, left=3cm, right=3cm]{geometry}

\usepackage{subfigure}

\newcommand{\HH}{\ensuremath{\mathbf{H}}}

\newcommand{\U}{\ensuremath{\mathbf{U}}}

\newcommand{\W}{\ensuremath{\mathbf{W}}}

\renewcommand{\c}{\ensuremath{\mathbf{c}}}

\newcommand{\f}{\ensuremath{\mathbf{f}}}
\newcommand{\g}{\ensuremath{\mathbf{g}}}
\newcommand{\h}{\ensuremath{\mathbf{h}}}

\newcommand{\rr}{\ensuremath{\mathbf{r}}}
\newcommand{\sss}{\ensuremath{\mathbf{s}}}  

\newcommand{\vv}{\ensuremath{\mathbf{v}}}

\newcommand{\x}{\ensuremath{\mathbf{x}}}
\newcommand{\y}{\ensuremath{\mathbf{y}}}
\newcommand{\z}{\ensuremath{\mathbf{z}}}

{%
\begin{list}{#1}{
\vspace{-\topsep}
\vspace{-\partopsep}
\setlength{\itemindent}{0cm}
\setlength{\rightmargin}{0cm}
\setlength{\listparindent}{0cm}
\settowidth{\labelwidth}{#1}
\setlength{\leftmargin}{\labelwidth}
\addtolength{\leftmargin}{\labelsep}
\setlength{\itemsep}{0cm}
}%
}%
{%
\end{list}
\vspace{-\topsep}
\vspace{-\partopsep}
}

{\begin{enumerate}%
}%
{\end{enumerate}}

\aclfinalcopy

\title{Deep Neural Machine Translation with Linear Associative Unit}
\author{Mingxuan Wang$^1$ \ Zhengdong Lu$^2$ \
         Jie Zhou$^2$ \  Qun Liu$^{4,5}$ \\
         $^1$Mobile Internet Group, Tencent Technology Co., Ltd\\
         {\tt xuanswang@tencent.com}\\
        $^2$DeeplyCurious.ai\\
        $^3$ Insititute of Deep Learning Research, Baidu Co., Ltd \\
        $^4$ Institute of Computing Technology, Chinese Academy of Sciences\\
        $^5$ADAPT Centre, School of Computing, Dublin City University\\
}
\date{}

\begin{document}
\maketitle
\begin{abstract}
Deep Neural Networks (DNNs) have provably enhanced the
state-of-the-art Neural  Machine Translation (NMT) with
 their capability in modeling complex functions and capturing
  complex linguistic structures.
  However NMT systems with deep architecture in their encoder or
  decoder RNNs often suffer from severe gradient diffusion
  due to the non-linear recurrent activations, which often
  make the optimization much more difficult.
   To address this problem we propose  novel linear
   associative units (LAU)  to reduce the gradient
    propagation length inside the recurrent unit.
    Different from conventional
    approaches (LSTM unit and GRU),
   LAUs utilizes linear associative connections
   between input and
   output of the recurrent unit,
   which allows unimpeded information flow through both
    space and time direction.  The model is quite simple,
     but it is surprisingly effective. Our empirical
     study on Chinese-English translation shows that our
     model with proper configuration can improve
      by 11.7 BLEU upon Groundhog and the best
      reported  results in the same setting.
      On WMT14 English-German task and a larger WMT14
       English-French task, our
 model achieves comparable results with the state-of-the-art.
\end{abstract}

\section{Introduction}
Neural Machine Translation (NMT)
is an end-to-end learning approach to machine  translation
which has recently shown
promising results on multiple language
pairs~\cite{luong2015effective,shen2015minimum,wu2016google,zhang2016variational,tu2016modeling,zhang2016exploiting,jean-EtAl:2015:ACL-IJCNLP,DBLP:journals/corr/MengLTLL15}.
Unlike conventional Statistical Machine
Translation (SMT) systems~\cite{koehn2003statistical,chiang2005hierarchical,liu2006tree,xiong2006maximum,mi2008forest} which consist of multiple
separately tuned components,
NMT aims at building upon a single and large neural
network to directly map input
text to associated output text.
Typical NMT models consists of two recurrent neural networks (RNNs),
an encoder to read and encode the input text into a distributed
representation and a decoder to generate translated text
conditioned on the input representation~\cite{sutskever2014sequence,RNNsearch}.

Driven by the breakthrough achieved in computer vision~\cite{he2015deep,srivastava2015training},
research in NMT has recently turned towards studying Deep Neural Networks (DNNs).
Wu et al.~\shortcite{wu2016google} and Zhou et al.~\shortcite{zhou2016deep} found that deep architectures in both the encoder and decoder are essential for capturing subtle irregularities in the source and target languages. However, training a deep neural network is not as simple as stacking layers.
Optimization often becomes increasingly difficult with more layers.
One reasonable explanation is the notorious
problem of vanishing/exploding gradients which
was first studied in the context of vanilla RNNs~\cite{pascanu2013difficulty}.
Most prevalent approaches to solve this problem
rely on short-cut connections between adjacent layers
such as residual or fast-forward connections~\cite{he2015deep,srivastava2015training,zhou2016deep}.
Different from previous work, we choose to reduce the gradient path inside the recurrent units
and
propose a novel Linear Associative Unit (LAU)
which creates a fusion of both linear and non-linear transformations of the input.
Through this design, information can flow across several steps both in time and in space with
little attenuation.
The mechanism makes it easy to train
deep stack RNNs which can efficiently
capture the complex inherent structures of sentences for NMT.
Based on LAUs, we also propose a NMT model , called \textsc{DeepLAU}, with deep architecture in both the encoder and decoder.

Although \textsc{DeepLAU} is fairly simple,
it gives remarkable empirical results.
On the NIST Chinese-English task,
\textsc{DeepLAU} with proper settings yields the best reported result
and also a 4.9 BLEU improvement over a strong NMT
baseline with most known techniques (e.g, dropout)
incorporated.
On WMT English-German and English-French tasks,
it also achieves performance superior or
comparable to the state-of-the-art.

\section{Neural machine translation}

A typical neural machine translation system is a single and large neural network
which directly models the conditional probability $p(\y|\x)$ of translating a source
sentence $\x = \{x_1, x_2, \cdots, x_{T_x} \}$ to a target sentence $\y = \{y_1, y_2,\cdots,
y_{T_y} \}$.

Attention-based NMT, with RNNsearch as its most popular representative,
generalizes the conventional notion of encoder-decoder in using an array of vectors to represent the
source sentence and dynamically addressing the relevant segments of them during decoding.
The process can be explicitly split into an encoding part, a decoding part and an attention mechanism.
The model first encodes the source sentence $\x$ into a sequence of vectors $\c=\{h_1,h_2,\cdots,h_{T_x}\}$.
In general, $h_i$ is the annotation of $x_i$ from a bi-directional RNN which contains
information about the whole sentence with a strong focus on the parts of $x_i$.
Then, the RNNsearch model decodes and generates the target translation $\y$ based on the context $\c$
and the partial traslated sequence $\y_{<t}$ by maximizing the probability of $p(y_i|y_{<i},\c)$.
In the attention model, $\c$ is dynamically obtained according to the contribution of the source annotation made to the word prediction.
This is called automatic alignment~\cite{RNNsearch} or attention mechanism~\cite{luong2015effective},
but it is essentially reading with content-based addressing defined in~\cite{NTM}. With this addressing strategy the
decoder can attend to the source representation that is most relevant to the stage of decoding.

Deep neural models have recently achieved a great success in a wide range of
problems. In computer vision, models with more than $100$ convolutional layers
have outperformed shallow ones by a big margin on a series of image tasks~\cite{he2015deep,srivastava2015training}.
Following  similar ideas of building deep CNNs,
some promising improvements have also been achieved on building deep NMT systems.
Zhou et al.~\shortcite{zhou2016deep} proposed a new type of linear connections between
adjacent layers to simplify the training of deeply stacked RNNs.
Similarly, Wu et al.~\shortcite{wu2016google} introduced residual connections to their deep
neural machine translation system and achieve great improvements.
However the optimization of deep RNNs is still an open problem
due to the massive
recurrent computation which makes the gradient propagation path
extremely tortuous.

\section{Model Description}

In this section, we discuss Linear Associative Unit (LAU) to ease the training of deep stack of RNNs.
Based on this idea, we further propose  \textsc{DeepLAU}, a neural machine translation model with a deep encoder and decoder.

\subsection{Recurrent Layers}
A recurrent neural network ~\cite{williams1989learning} is a
class of neural network that has recurrent connections and a state (or its more sophisticated memory-like extension). The past
information is built up through the recurrent connections.
This makes RNN applicable for sequential
prediction tasks of arbitrary length. Given a
sequence of vectors $\x = \{\x_1, \x_2, \cdots, \x_{T} \}$ as input, a
standard RNN computes the sequence hidden states $\h = \{\h_1, \h_2, \cdots, \h_{T} \}$
by iterating the following equation
from $t=1$ to $t=T$:
\begin{equation}
\h_t = \phi(\x_t, \h_{t-1})
\end{equation}
$\phi$ is usually a nonlinear function such as composition of a logistic sigmoid with an affine transformation.

\subsection{Gated Recurrent Unit}
It is difficult to train RNNs
to capture long-term dependencies because the gradients tend to either vanish (most of the time) or
explode. The effect of long-term dependencies is dropped exponentially with respect to the gradient propagation length.
The problem was explored in depth by ~\cite{hochreiter1997long,pascanu2013difficulty}.
A successful approach is to design a more sophisticated
activation function than a usual activation function
consisting of gating functions to control the information flow and reduce the propagation path.
There is a long thread of work aiming to solve this problem,
with the long short-term memory units (LSTM)  being the most
salient examples and gated recurrent unit (GRU) being the most recent one~\cite{hochreiter1997long,GRU}.
RNNs employing either
of these recurrent units have been shown to perform well in tasks that require capturing long-term
dependencies.

GRU  can be viewed as a slightly more dramatic variation on  LSTM with fewer parameters.
The activation function is armed with two specifically designed gates called update and reset gates
to control the flow of information
inside each hidden unit. Each hidden state at time-step $t$ is computed as follows
\begin{equation}
\h_t = (1-\z_t)\odot\h_{t-1} + \z_t\odot\tilde\h_t
\end{equation}
where $\odot$ is an element-wise product, $\z_t$ is the update gate,
and $\tilde \h_t$ is the candidate activation.
\begin{equation}
\tilde\h_t = \tanh(\W_{xh}\x_t + \W_{hh}(\rr_t\odot\h_{t-1}))
\end{equation}
where $\rr_t$ is the reset gate. Both reset and update gates are computed as :
\begin{align}
\rr_t ={}& \sigma(\W_{xr}\x_t+\W_{hr}\h_{t-1})\label{eq:gate1}\\
\z_t ={}& \sigma(\W_{xz}\x_t+\W_{hz}\h_{t-1})\label{eq:gate2}
\end{align}
This procedure of taking a linear sum between the existing state and the newly computed state is
similar to the LSTM unit.

\subsection{Linear Associative Unit}
GRU can actually be viewed as a non-linear activation function with gating mechanism.
Here we propose LAU which extends GRU by having an additional
linear transformation of the input in its dynamics. More formally, the state update function becomes
\begin{equation}\label{eq:lau}
\begin{split}
\h_t =  &((1-\z_t)\odot \h_{t-1}+\z_t\odot \tilde \h_t)\odot(1-\g_t) \\
 & + \g_t\odot \HH(\x_t).
\end{split}
\end{equation}
Here  the updated $\h_t$ has three sources: 1) the direct transfer from previous state $\h_{t-1}$,  2) the candidate update  $\tilde \h_t $, and 3) a direct contribution from the input $\HH(\x_t)$.
 More specifically, $\tilde \h_t$ contains the nonlinear information of the input and the previous hidden state.
\begin{equation}
\tilde \h_t = \tanh(\f_t\odot(\W_{xh}\x_t)+\rr_t \odot (\W_{hh}\h_{t-1})),
\end{equation}
where $\f_t$ and $\rr_t$ express how much of the nonlinear abstraction
are produced by the input $\x_t$ and previous hidden state $\h_t$, respectively.
For simplicity, we set $\f_t=1-\rr_t$ in this paper and find that this works
well in our experiments.  The term $\HH(\x_t)$ is usually an affine linear
transformation of the input $\x_t$ to mach the dimensions of $\h_t$,
where $\HH(\x_t) = \W_{x} x_t$.   The associated term
$\g_t$  (the input gate) decides how much of the linear transformation of the
input is carried  to the hidden state and then the output. The gating function $\rr_t$ (reset gate)
and $\z_t$ (update gate) are computed following Equation~(\ref{eq:gate1}) and~(\ref{eq:gate2})  while $\g_t$ is computed as
\begin{equation}
\g_t = \sigma(\W_{xg}\x_t+\W_{hg}\h_{t-1}).
\end{equation}
The term $ \g_t\odot \HH(\x_t)$ therefore offers a direct way for
input $\x_t$ to go to later hidden layers, which can eventually lead to a path to the output layer when applied recursively.
This mechanism is potentially very useful for translation where the input, no matter whether it is the source word or the attentive reading (context),
 should sometimes be directly carried to the next stage of
 processing without any substantial composition or nonlinear transformation.  To understand this, imagine we want to translate a English sentence containing a  relative rare entity name such as  ``Bahrain" to Chinese:  LAU is potentially able to retain the embedding of this word in its hidden state, which will  otherwise be prone to serious distortion due to the scarcity of training instances for it.

\subsection{\textsc{DeepLAU}}
\begin{figure}[h!]
\begin{center}
\includegraphics[width=0.49\textwidth]{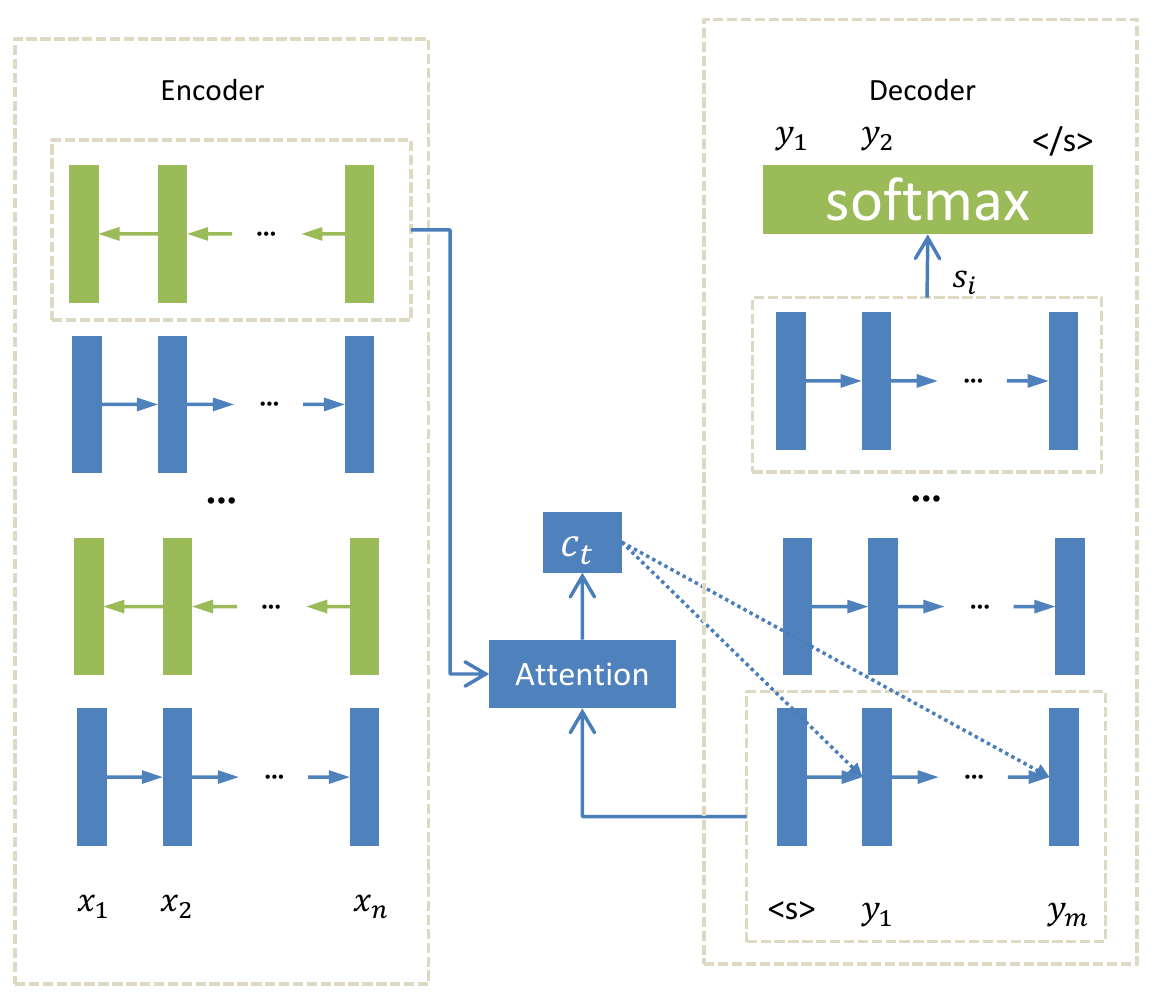}
\end{center}
\caption{\textsc{DeepLAU}: a neural machine translation
model with deep encoder and decoder.}\label{f:overall}
\end{figure}
Graves et al.~\shortcite{graves2013generating} explored the advantages of deep RNNs for handwriting recognition and text generation.
There are multiple ways of combining one layer of RNN with another. Pascanu et al.~\shortcite{pascanu2013construct} introduced
Deep Transition RNNs with Skip connections (DT(S)-RNNs).
Kalchbrenner et al.~\shortcite{kalchbrenner2015grid} proposed to make a full connection of all the RNN hidden layers.
In this work
we employ vertical stacking where only the output of the
previous layer of RNN is fed to the current layer as input.
The input at recurrent layer
$\ell$ (denoted as $\x_t^{\ell}$) is exactly the output
of the same time step at layer $\ell-1$ (denoted as $\h_t^{\ell-1}$).
Additionally, in order to learn more temporal
dependencies, the sequences can be processed in
different directions.
More formally, given an input sequence $\x = (\x_1,...,\x_T) $, the output on layer $\ell$ is
\begin{equation}\label{eq:dbrnn}
\h_t^{(\ell)} =\left\{\begin{array}{ll}
\x_t, & \ell = 1 \\
\phi^{\ell}(\h_{t+d}^{(\ell)},\h_{t}^{(\ell-1)}), & \ell>1
\end{array}\right.
\end{equation}
where
\begin{itemize}
\item $\h_t^{(\ell)}$ gives the output of layer $\ell$ at location $t$.
\item $\phi$ is a recurrent function and we choose LAUs in this work.
\item The directions are marked by a direction term $d \in \{-1, 1\}$.
If we fixed $d$ to $-1$, the input will be processed
in forward direction, otherwise backward direction.
\end{itemize}

The deep architecture of \textsc{DeepLAU}, as  shown in Figure~\ref{f:overall}, consists of three parts:  a stacked LAU-based encoder,  a stacked LAU-based decoder and an improved attention model.
\paragraph{Encoder}
One shortcoming of conventional RNNs is that they are only able to make use of previous context.
In machine translation, where whole source utterances are transcribed at once,
there is no reason not to exploit future context as well.
Thus bi-directional RNNs are proposed to integrate information from the past and the future.
The typical bidirectional approach processes the raw input in backward and forward direction with two separate  layers, and then concatenates them together.
Following Zhou et al.~\shortcite{zhou2016deep}, we choose another bidirectional approach to process
the sequence in order to learn more temporal dependencies in this work.
Specifically, an RNN layer processes the input sequence in forward direction.
The output of this layer is taken by an upper RNN layer as input, processed in reverse direction.
Formally, following Equation~(\ref{eq:dbrnn}), we set $d=(-1)^\ell$.
This approach can easily build a deeper network
with the same number of parameters compared to the classical approach.
The final encoder consists of $L_\text{enc}$ layers and produces the output
$\h^{L_\text{enc}}$ to compute the conditional input $\c$ to the decoder.

\paragraph{Attention Model}
The alignment model $\alpha_{t,j}$ scores how well the output
at position $t$ matches the inputs around position $j$ based on $\sss_{t-1}^1$ and $\h_j^{L_\text{enc}}$ where $\h_j^{L_\text{enc}}$ is the top-most layer of the encoder at step $j$
and $\sss_{t-1}^1$ is the first layer
of decoder at step $t-1$.
It is intuitively beneficial to exploit the information of $y_{t-1}$ when reading from the source sentence representation, which is missing from the implementation of attention-based NMT in~\cite{RNNsearch}.
In this work, we build a more effective alignment path by feeding both the previous hidden state $\sss_{t-1}^1$
and the context word $y_{t-1}$ to the attention model, inspired by the recent implementation of attention-based NMT\footnote{\url{github.com/nyu-dl/dl4mt-tutorial/tree/master/session2}}.
The conditional input $\c_j$ is a weighted sum of attention score $\alpha_{t,j}$ and encoder output $\h^{L_\text{enc}}$.
Formally, the calculation of $\c_j$ is
\begin{equation}\label{eq:att}
\c_j = \sum_{t=1}^{t=L_x} \alpha_{t,j}\h_t^{L_\text{enc}}
\end{equation}
where
\begin{equation}
\begin{aligned}
e_{t,j}={}& \vv_a^T\sigma(\W_a\sss_{t-1}^{1}+\U_a\h_j^{L_\text{enc}} + \W_y{\y_{t-1}})\\
\alpha_{t,j} ={}& \text{softmax}(e_{t,j}).
\end{aligned}
\end{equation}
$\sigma$ is a nonlinear function with the information
of $y_{t-1}$ (its word embedding being $\y_{t-1}$) added.
In our preliminary experiments, we found that GRU works slightly better than tanh function, but we chose the latter for simplicity.
\paragraph{Decoder}
The decoder follows Equation~(\ref{eq:dbrnn}) with fixed direction term $d=-1$.
At the first layer, we use
the following input:
$$\x_t=\left[ \c_t, \y_{t-1}\right]$$
where
$\y_{t-1}$ is the target word embedding at time step $t$,
$\c_t$ is dynamically obtained follows Equation~(\ref{eq:att}).
There are $L_\text{dec}$ layers of RNNs armed with LAUs in the decoder.
At inference stage, we only utilize the top-most hidden states
$\sss^{L_\text{dec}}$ to make the final prediction with a softmax layer:
\vspace{-10pt}
\begin{equation}
p(y_i|y_{<i},\x) = \text{softmax}(\W_o \sss_i^{L_\text{dec}})\vspace{-10pt}
\end{equation}.

\section{Experiments}
\subsection{Setup}
We mainly evaluated our approaches on the widely used NIST Chinese-English
translation task.
In order
to show the usefulness of our approaches, we also provide results on other two
translation
tasks: English-French, English-German.
The evaluation metric is BLEU\footnote{
For Chinese-English task, we apply case-insensitive NIST BLEU.
For other tasks, we tokenized the reference and evaluated the
performance with \emph{multi-bleu.pl}.
The metrics are exactly the same as in previous work.}
\cite{papineni2002bleu}.

For Chinese-English, our training data consists of $1.25$M sentence
pairs extracted from LDC corpora\footnote{The corpora include LDC2002E18,
LDC2003E07, LDC2003E14, Hansards portion of LDC2004T07, LDC2004T08 and LDC2005T06.},
with $27.9$M Chinese words and $34.5$M English words respectively.
We choose NIST 2002 (MT02) dataset as our development set, and the NIST 2003 (MT03),
2004 (MT04) 2005 (MT05) and 2006 (MT06) datasets as our test sets.

For English-German, to compare with the
results reported by previous work~\cite{luong2015effective,zhou2016deep,jean-EtAl:2015:ACL-IJCNLP}, we used the same subset
of the WMT 2014 training corpus that contains
4.5M sentence pairs with 91M English words and
87M German words. The concatenation of news-test
2012 and news-test 2013 is used as the validation
set and news-test 2014 as the test set.

To evaluate at scale, we also report the results of
English-French.
To compare with the results
reported by previous work on end-to-end NMT
~\cite{sutskever2014sequence,RNNsearch,jean-EtAl:2015:ACL-IJCNLP,luong2014addressing,zhou2016deep},
we used
the same subset of the WMT 2014 training corpus
that contains 12M sentence pairs with 304M
English words and 348M French words. The concatenation
of news-test 2012 and news-test 2013
serves as the validation set and news-test 2014 as
the test set.

\subsection{Training details}
\begin{table*}[!h]
\begin{center}
\begin{tabular}{l|cccc|c}

SYSTEM & MT03 & MT04 & MT05 & MT06 & AVE.\\
\hline
\hline
\multicolumn{6}{c}{Existing systems} \\
\hline
Moses & $31.61$ & $33.48$ & $30.75$ & $30.85$ & $31.67$ \\
Groundhog & $31.92$ & $34.09$ & $31.56$ & $31.12$ & $32.17$ \\
\textsc{coverage}& $34.49$ & $38.34$ & $34.91$ & $34.25$ & $35.49$ \\
\textsc{MemDec} & $36.16$ & $39.81$ & $35.91$ & $35.98$ & $36.95$ \\
\hline
\multicolumn{6}{c}{Our deep NMT systems} \\
\hline
\textsc{DeepGRU} & $33.21$ & $36.76$ & $33.05$ & $33.30$ & $34.08$ \\
\textsc{DeepLAU} & $\textbf{39.35}$ & $\textbf{41.15}$ & $\textbf{38.07}$ & $\textbf{37.29}$ & $\textbf{38.97}$ \\
\hline
\textsc{DeepLAU} +Ensemble + PosUnk & $42.21$ & $43.85$ & $44.75$ & $42.58$ & $43.35$ \\

\end{tabular}

\end{center}
\caption{\label{tab:CEEXP} Case-insensitive BLEU scores on Chinese-English translation.
}
\end{table*}

Our training procedure and hyper parameter
choices are similar to those used by~\cite{RNNsearch}.
In more details,
we limit the source and target vocabularies
to the most frequent $30K$ words in both Chinese-English and
English-French.
For English-German, we set the source and target vocabularies size to $120K$ and
$80K$, respectively.

For all  experiments, the dimensions of  word embeddings and recurrent hidden states
are both set to $512$. The dimension of $c_t$ is also of size $512$.
Note that our network is more narrow than most previous work
where  hidden states of dimmention $1024$ is used.
we initialize parameters by sampling each element from the Gaussian distribution with mean $0$
and variance $0.04^2$.

Parameter optimization is performed using stochastic gradient descent.
Adadelta~\cite{zeiler2012adadelta} is used to automatically
adapt the learning rate of each parameter ($\epsilon=10^{-6}$
and $\rho=0.95$).
To avoid gradient explosion, the gradients of the cost function which had $\ell_2$
 norm larger than a predefined threshold $\tau$ were normalized to the threshold~\cite{pascanu2013construct}.
 We set $\tau$ to $1.0$  at the beginning and halve the threshold until the BLEU score does not
change much on the development set.
Each SGD is a mini-batch of $128$ examples.
We train our NMT model with the sentences of
length up to $80$ words in the training data,
while for the Moses system we use the full training data.
Translations are generated by a beam search and log-likelihood
scores are normalized by sentence length.
We use a beam width of $10$ in all the experiments.
Dropout was also applied on the output layer to avoid over-fitting.
The dropout rate is set to $0.5$.
Except when otherwise mentioned,  NMT systems are have 4 layers encoders and 4 layers decoders.
\subsection{Results on Chinese-English Translation}

Table~\ref{tab:CEEXP} shows BLEU scores on Chinese-English datasets.
Clearly \textsc{DeepLAU} leads to a remarkable improvement over their competitors.
Compared to \textsc{DeepGRU},
\textsc{DeepLAU} is $+4.89$ BLEU score
higher on average four test sets,
showing the modeling power gained from the liner associative
connections.
We suggest it is because LAUs apply adaptive gate function conditioned on the input
which make it able to automatically decide how much linear information should be transferred
to the next step.

To show the power of \textsc{DeepLAU}, we also make a comparison with previous work.
Our best single model outperforms both a phrased-based MT system (Moses) as well as an open
source attention-based NMT system (Groundhog)
by $+7.3$ and $+6.8$ BLEU points respectively on average.
The result is also better than some
other state-of-the-art variants of attention-based NMT mode
with big margins.
After PosUnk and ensemble, \textsc{DeepLAU} seizes another notable gain of
$+4.38$ BLEU and outperform Moses by $+11.68$ BLEU.
\subsection{Results on English-German Translation}
\begin{table*}[!h]
\begin{center}
\begin{tabular}{l|l|c|c}

SYSTEM & Architecture & Voc. & BLEU \\
\hline
\hline
\multicolumn{4}{c}{Existing systems} \\
\hline
Buck et al. ~\shortcite{buck2014n} & Winning WMT’14 system – phrase-based + large LM &-& $20.7$ \\
\hline
Jean et al. ~\shortcite{jean-EtAl:2015:ACL-IJCNLP} & gated RNN with search + LV + PosUnk & $500K$ & $19.4$ \\
Luong et al. ~\shortcite{luong2015effective} & LSTM with $4$ layers + dropout + local att. + PosUnk & $80K$ & $20.9$ \\
Shen et al. ~\shortcite{shen2015minimum} & gated RNN with search + PosUnk + MRT & $80K$ & $20.5$ \\
Zhou et al. ~\shortcite{zhou2016deep} & LSTM with $16$ layers + F-F connections& $80K$ & $20.6$ \\
Wu et al. ~\shortcite{wu2016google} & LSTM with $8$ laysrs + RL-refined Word & $80K$ & 23.1 \\
Wu et al. ~\shortcite{wu2016google} & LSTM with $8$ laysrs + RL-refined WPM-32K & - & 24.6 \\
Wu et al. ~\shortcite{wu2016google} &LSTM with $8$ laysrs + RL-refined WPM-32K + Ensemble & - & 26.3 \\
\hline
\multicolumn{4}{c}{Our deep NMT systems} \\
\hline
this work & \textsc{DeepLAU} & $80K$ & $22.1(\pm0.3)$ \\
this work &\textsc{DeepLAU} + PosUnk& $80K$ & $23.8(\pm0.3)$ \\
this work & \textsc{DeepLAU} + PosUnk + Ensemble $8$ models & $80K$ & $26.1$ \\

\end{tabular}

\end{center}
\caption{\label{tab:GEEXP} Case-sensitive BLEU scores on German-English translation.
}
\end{table*}
The results on English-German translation
are presented in Table~\ref{tab:GEEXP}.
We compare our NMT systems with various other systems
including the winning system in WMT’14~\cite{buck2014n}, a phrase-based system whose language
models were trained on a huge monolingual text,
the Common Crawl corpus. For end-to-end NMT
systems, to the best of our knowledge,
Wu et al.~\shortcite{wu2016google} is currently the SOTA system and about $4$ BLEU points on top of previously best
reported results even though Zhou
et al.~\shortcite{zhou2016deep} used a much deeper neural network\footnote{It is also worth mentioning that
the result reported by Zhou
et al.~\shortcite{zhou2016deep} does not include PosUnk, and this comparison is not fair enough.}.

Following Wu et al.~\shortcite{wu2016google}, the BLEU score represents the
averaged score of 8 models we trained.
Our approach achieves comparable results
with SOTA system.
As can be seen from the Table~\ref{tab:GEEXP}, DeepLAU performs better than
the word based model and even not much worse
than the  best wordpiece models achieved by Wu et al.~\shortcite{wu2016google}.
Note that \textsc{DeepLAU} are simple and easy to implement,
as opposed to previous models reported in Wu et al.~\shortcite{wu2016google},
which dependends on some external techniques to achieve their
best performance, such as their introduction of length normalization,
coverage penalty, fine-tuning and the RL-refined model.
\subsection{Results on English-French Translation}
\begin{table}[!h]
\begin{center}
\begin{tabular}{l|c}

SYSTEM & BLEU\\
\hline
\hline
Enc-Dec \cite{luong2014addressing} & $30.4$ \\
RNNsearch \cite{RNNsearch} & $28.5$ \\
RNNsearch-LV \cite{jean-EtAl:2015:ACL-IJCNLP} & $32.7$ \\
Deep-Att \cite{zhou2016deep}& $35.9$ \\
\hline
\textsc{DeepLAU} & $35.1$ \\
\end{tabular}
\end{center}
\caption{\label{tab:EFEXP}
English-to-French task: BLEU scores of single
neural models.
}
\end{table}
To evaluate at scale, we also show the results on an English-French task
with $12M$ sentence pairs and $30K$ vocabulary
in Table~\ref{tab:EFEXP}.
Luong et al.~\shortcite{luong2014addressing} achieves BLEU score of $30.4$ with a six layers
deep Encoder-Decoder model.
The two attention models, RNNSearch and RNNsearch-LV achieve BLEU
scores of $28.5$ and $32.7$ respectively.
The previous best single NMT Deep-Att model with an $18$ layers
encoder and  $7$ layers decoder achieves BLEU score of $35.9$.
For \textsc{DeepLAU}, we obtain the BLEU score of $35.1$ with a $4$ layers
encoder and $4$ layers decoder, which is
on par with the SOTA system in terms of BLEU.
Note that Zhou et al. ~\shortcite{zhou2016deep} utilize a much larger depth as well as
external alignment model and extensive
regularization to achieve their best results.

\subsection{Analysis}
Then we will study the main factors that influence our results
on NIST Chinese-English translation task.
We also compare our approach with two SOTA
topologies which were used in building deep NMT systems.

\begin{itemize}
\item {
Residual Networks (ResNet) are among the pioneering
works ~\cite{szegedy2016inception,he2016deep} that utilize extra identity connections to
enhance information flow such that very deep neural
networks can be effectively optimized.
Share the similar idea, Wu et al.~\shortcite{wu2016google} introduced to leverage residual connections to train
deep RNNs.
}
\item{Fast Forward (F-F) connections were proposed to reduce the propagation
path length which is the pioneer work to simplify the training of deep NMT model ~\cite{zhou2016deep}.
The work can be viewed as a parametric ResNet with
short cut connections between adjacent layers.
The procedure takes a linear sum between the input and the newly computed
state.
}
\end{itemize}
\paragraph{LAU vs. GRU}
\begin{table}[!h]
\begin{center}
\begin{tabular}{l|cc|c}

SYSTEM & ($L_\text{enc}$,$L_\text{Dec})$ & width & AVE.\\
\hline
\hline
\small{1} \textsc{DeepGRU} & ($2$,$1$) & $512$ & $33.59$ \\
\small{2} \textsc{DeepGRU} & ($2$,$2$) & $1024$ & $34.68$ \\
\small{3} \textsc{DeepGRU} & ($2$,$2$) & $512$ & $34.91$ \\
\small{4} \textsc{DeepGRU} & ($4$,$4$) & $512$ & $34.08$ \\
\hline
\small{5} 4+ResNet & ($4$,$4$) &$512$ & $36.40$ \\
\small{6} 4+F-F& ($4$,$4$) &$512$ & $37.62$ \\
\hline
\small{7} \textsc{DeepLAU} & ($2$,$2$) & $512$ & $37.65$ \\
\small{8} \textsc{DeepLAU} & ($4$,$4$) & $512$ & $38.97$ \\
\small{9} \textsc{DeepLAU} & ($8$,$6$) & $512$ & $39.01$ \\
\small{10} \textsc{DeepLAU} & ($8$,$6$) & $256$ & $38.91$ \\
\end{tabular}

\end{center}
\caption{\label{tab:AEXP}
BLEU scores of \textsc{DeepLAU} and \textsc{DeepGRU}
 with different model sizes.
}
\end{table}
Table~\ref{tab:AEXP} shows the effect of the novel LAU.
By comparing row $3$ to row $7$, we see that
when $L_\text{Enc}$ and $L_\text{Dec}$ are set to $2$,
the average BLEU scores achieved by \textsc{DeepGRU} and \textsc{DeepLAU} are $34.68$ and $37.65$, respectively.
LAU can bring an improvement of $2.97$ in terms of BLEU.
After increasing the model depth to $4$ (row $4$ and row $6$), the improvement is enlarged to $4.91$.
When \textsc{DeepGRU} is trained with larger depth (say, $4$), the training becomes more difficult and
the performance falls behind its shallow partner.
While for \textsc{DeepLAU}, as can be see in row $9$, with increasing the depth even to $L_\text{Enc}=8$ and
$L_\text{Dec}=6$ we can still obtain  growth by $0.04$ BLEU score.
Compared to previous short-cut connection methods (row $5$ and row $6$),
The LAU still achieve meaningful improvements over F-F connections and Residual connections by
$+1.35$ and $+2.57$ BLEU points respectively.

\textsc{DeepLAU} introduces more parameters than \textsc{DeepGRU}.
In order to figure out the
effect of \textsc{DeepLAU} comparing models with the same parameter
size, we increase the hidden size of \textsc{DeepGRU} model.
Row $3$ shows that, after using a twice larger GRU layer,
the BLEU score is $34.68$, which is still worse than the corresponding
\textsc{DeepLAU} model with fewer parameters.

\paragraph{Depth vs. Width}
Next we will study the model size.
In Table~\ref{tab:AEXP}, starting from $L_\text{Enc}=2$ and $L_\text{Dec}=2$ and
gradually increasing the model depth, we can achieve substantial
improvements in terms of BLEU.
With $L_\text{Enc}=8$ and $L_\text{Dec}=6$, our \textsc{DeepLAU} model
yields the best BLEU score.
We tried to increase the model depth with the same hidden size but failed to
see further improvements.

We then tried to increase the hidden size.
By comparing row $2$ and row $3$, we find the improvements is relative small
with a wider hidden size.
It is also worth mentioning that a deep and thin
network with fewer parameters
can still achieve comparable results with
its shallow partner.
This suggests that depth plays a more important role in increasing
the complexity of neural networks than width and our deliberately designed
LAU benefit from the optimizing of such a deep model.
\paragraph{About Length}
\begin{figure}[h!]
\begin{center}
\includegraphics[width=0.50\textwidth]{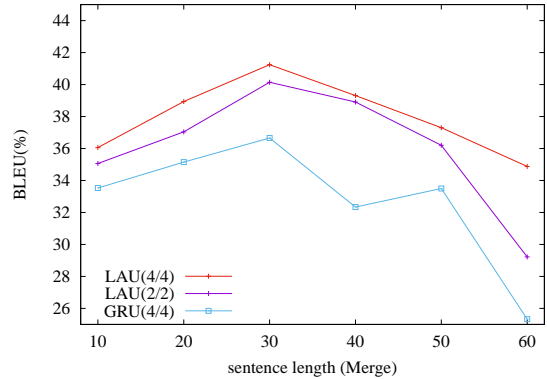}
\label{f:long}
\end{center}\vspace{-10pt}
\caption{The BLEU scores of generated translations on
the merged four test sets with respect to the lengths of
source sentences.}
\end{figure}
A more detailed comparison between \textsc{DeepLAU} (4 layers encoder and 4 layers decoder),
\textsc{DeepLAU}(2 layer encoder and 2 layer decoder) and \textsc{DeepGRU} (4 layers encoder and 4 layers decoder),
suggest that with deep architectures are essential to the superior performance of
our system.
In particular, we test the BLEU scores on sentences longer than $\{10, 20, 30, 40, 50,60\}$ on the
merged test set.
Clearly, in all curves, performance degrades with increased sentence length.
However, \textsc{DeepLAU} models yield consistently higher BLEU scores than the \textsc{DeepGRU} model on longer sentences.
These observations are consistent with our intuition that very deep RNN model is especially good at
modeling the nested latent structures on relatively complicated sentences and LAU plays an important role
on optimizing such a complex deep model.

\section{Conclusion}
We propose a Linear Associative Unit (LAU)
which makes a fusion of both linear and non-linear transformation inside the recurrent unit.
On this way, gradients decay
much slower compared to the standard deep networks which enable us
to build a deep neural network for machine translation.
Our empirical study shows that it can significantly
improve the performance of NMT.

\section{acknowledge}
We sincerely thank the anonymous reviewers for their thorough
reviewing and valuable suggestions.
Wang's work is partially supported by National Science Foundation for  Deep
Semantics Based Uighur to Chinese Machine Translation (ID 61662077).
Qun Liu's work is partially supported by Science Foundation Ireland in the ADAPT Centre for Digital Content Technology (www.adaptcentre.ie) at Dublin City University funded under the SFI Research Centres Programme (Grant 13/RC/2106) co-funded under the European Regional Development Fund.
\bibliography{deep_nmt}
\bibliographystyle{acl_natbib}
\end{document}